\pdfoutput=1
\documentclass[11pt]{article}

\usepackage{ACL2023}

\usepackage{times}
\usepackage{latexsym}

\usepackage[T2A,T1]{fontenc}
\usepackage[utf8]{inputenc}

\usepackage{microtype}

\usepackage{inconsolata}

\usepackage{color, soul}  % for \hl

\usepackage{amsmath}
\usepackage{amssymb}
\usepackage{breqn}
\usepackage{upgreek}

\usepackage{graphicx}
\usepackage{subcaption}
\usepackage{cleveref}
\usepackage{mathtools}

\usepackage{booktabs}
\newcommand{\bftab}{\fontseries{b}\selectfont}
\title{End-to-End Argument Mining over Varying Rhetorical Structures}

\author{Elena Chistova \\
  FRC CSC RAS, Moscow, Russia \\
  \texttt{chistova@isa.ru} 
  }

\begin{document}
\maketitle
\begin{abstract}
%The study highlights that using a single rhetorical annotation can result in biased results when analyzing the connections between discourse and argumentation. 
Rhetorical Structure Theory implies no single discourse interpretation of a text, and the limitations of RST parsers further exacerbate inconsistent parsing of similar structures. Therefore, it is important to take into account that the same argumentative structure can be found in semantically similar texts with varying rhetorical structures. In this work, the differences between paraphrases within the same argument scheme are evaluated from a rhetorical perspective. The study proposes a deep dependency parsing model to assess the connection between rhetorical and argument structures. The model utilizes rhetorical relations; RST structures of paraphrases serve as training data augmentations. The method allows for end-to-end argumentation analysis using a rhetorical tree instead of a word sequence. It is evaluated on the bilingual Microtexts corpus, and the first results on fully-fledged argument parsing for the Russian version of the corpus are reported. The results suggest that argument mining can benefit from multiple variants of discourse structure.\footnote{The code is available at \url{https://github.com/tchewik/e2e-microtexts}}
\end{abstract}

%%%%%%%%%%%%%%%%%%%%%%%%%%%%%%%%%%%%%%%%%
\section{Introduction}

The goal of argument mining is to automatically identify the premises, claims, and conclusions in an argument. Another field of NLP aiming to recognize structure in a complex text is discourse parsing. It involves identifying the author's point of view, the central idea, and the relations between discourse units. Rhetorical Structure Theory \citep{mann1988rhetorical} depicts text structure as a tree spanning the entire text, with rhetorical relations connecting adjacent text spans from elementary discourse units (EDUs) to paragraphs. 
Many efforts \citep{azar1999argumentative,villalba2012some,green2010representation,peldszus-stede-2016-rhetorical,stede-etal-2016-parallel,accuosto-saggion-2019-transferring} have been devoted to finding correlations between the two structure descriptions. The studies examine a single rhetorical parsing result or a single manual annotation for each text. However, the same argumentative structure can be found in semantically similar texts with varying rhetorical structures, especially when retrieved by automatic parsing. This must be taken into account when probing discourse against argumentation.

According to \citet{morey-etal-2017-much}, the human baseline score on the news-domain RST-DT \citep{carlson-etal-2001-building} benchmark is 55.0\% Parseval F1 for gold segmentation. An analyzer's inevitable mispredictions exacerbate inconsistent parsing of similar structures. Interpreting discourse accurately may require sophisticated skills, such as reasoning over general knowledge and assessing the subjective significance of particular statements. End-to-end discourse tree prediction recently achieved %46.6\% F1 on the RST-DT corpus \citep{nguyen-etal-2021-rst}.
50.1\% F1 on the RST-DT corpus \citep{liu-etal-2021-dmrst}. Discourse parsing is also significantly affected by domain shift. For an isolated subtask of RST-relation classification for pairs of adjacent EDUs in news, academic texts, TED talks, Reddit posts, and fiction (annotated in the GUM RST corpus \cite{zeldes2017gum}), \citet{atwell-etal-2022-change} report an averaged transfer error of 60\%. \citet{liu-zeldes-2023-cant} demonstrate that unlabeled RST tree construction performance degrades significantly when training on the WSJ-only RST-DT corpus and testing on the multidomain GUM. It degrades by $\sim 11$ points on average for spans only and by $\sim 16$ with nuclearities attached. These points are halved when testing only on the Wikinews-sourced \textit{news} part of GUM.

We argue that the analysis of the correlations between RST and argumentation is biased by the use of a single rhetorical annotation. These correlations can therefore be better assessed by using multiple rhetorical annotations of the same argumentative structures.
In this work, we propose a simple neural model, Discourse-driven Biaffine Parser (DBAP), to estimate the utility of labeled rhetorical structure for argument mining on short argumentative texts. We use the Argumentative Microtexts corpus proposed by \citet{peldszus2015annotated}. In this corpus, an argumentative text is seen as a hypothetical dialectical exchange between the author, who introduces and defends their claim, and their opponent. The argumentation can be represented by a graph with nodes corresponding to propositions expressed in textual segments, and edges indicating various supporting and attacking moves. We obtain two RST structures for each document by back translating over the parallel corpus of argument annotations. Then, we use the predicted RST structures in biaffine dependency parsing to estimate the general effect of rhetorical features.

To the best of our knowledge, this is the first end-to-end argument parser trained on a small corpus of Argumentative Microtexts and the first application using multiple versions of rhetorical structures to explore the relationship between discourse and argumentation. We also report the first results on fully-fledged argumentation mining for the Russian version of the corpus.

%%%%%%%%%%%%%%%%%%%%%%%%%%%%%%%%%%%%%%%%%
\section{Background and Related Work}
\label{sec:related_work}

A number of studies have examined the relationship between discourse and argumentation in monological texts. \citet{azar1999argumentative} suggests treating the five relations of the original RST as argumentative: \textsc{Motivation}, \textsc{Antithesis} and \textsc{Concession}, \textsc{Evidence}, and \textsc{Justify}. According to the hypothesis, one discourse unit is expected to influence the reader in relation to the other discourse unit.
Another investigation of the argumentativeness of rhetorical relations was carried out by \citet{villalba2012some}. The regularities in expressing persuasive arguments in the support function through certain cases of rhetorical relations \textsc{Elaboration}, \textsc{Justification}, \textsc{Restatement}, and \textsc{Comparison} and in the attack function through \textsc{Contrast} are demonstrated by a thorough analysis of online textual reviews.

\citet{green2010representation} combines some RST relations with the argumentative relations of \citet{toulmin1958uses} and \citet{walton2011refute} for a hybrid -- ArgRST -- manual annotation in a biomedical corpus of patient letters. In the later paper on the annotation of full-text biomedical research papers, \citet{green2015annotating} concludes that in a text of arbitrary genre, argumentation and discourse coherence should be represented separately. A hybrid representation of both schemes can also be achieved by annotating the rhetorical trees with communicative actions \citep{galitsky2018detecting} or enriching existing RST-dependency annotations with an argumentative structure layer \citep{accuosto-saggion-2019-transferring}.

The extended Microtexts corpus presented by \citet{stede-etal-2016-parallel} allows for the exploration of correlations between discourse and argumentation. It includes manual RST, PDTB, and Segmented Discourse annotation for 112 texts from the first version of the Microtexts corpus. 
They found, in particular, that 60\% of the argumentation arcs match those in RST; \textsc{Reason}, \textsc{Cause}, and \textsc{Evidence} RST relations are all most likely to match the support argumentation function; almost any RST relation can be found within the argumentative discourse unit (ADU).
\citet{peldszus-stede-2016-rhetorical} use the same manual RST annotations to train the argument parser. They construct a structure aligner and train evidence graph model \cite{peldszus-stede-2015-joint}, but using discourse rather than lexical features. Such features include the absolute and relative position of the segment in the text, whether the segment has incoming/outgoing RST edges, the number of edges, and the corresponding relations. For subgraphs of length~$ > 2$ also all chains of relations including this segment. The best performance is achieved when considering a subgraph of depth 3. RST parsing is first used to analyze arguments in Microtexts by \citet{hewett-etal-2019-utility}. The texts were analyzed with multiple earlier parsers, and the one proposed by \citet{feng-hirst-2014-linear} was chosen based on the manual evaluation of the results. The features used in the classifiers are the number of DUs of higher and lower levels; the same for the preceding and following DUs; the distance to the parent node; whether the segment is in a multinuclear relation. The proposed features insignificantly improved the argument analysis performance on the gold segmentation.

The earlier work examined an expert or early RST parser annotation of each document, while our work focuses on applying modern rhetorical parsers to explore the discourse variation in the short argumentative texts in English and Russian.

%%%%%%%%%%%%%%%%%%%%%%%%%%%%%%%%%%%%%%%%%
\section{Methods}
\label{sec:methods}

To analyze Argumentative Microtexts, we follow the classical Evidence Graphs approach of \citet{peldszus-stede-2015-joint}, where the argumentation graphs are directly converted into dependency trees.
However, unlike the Evidence Graphs method inferring the labeled argumentative structure from the results of complex cooperation between the structure, function, role, and central claim classifiers, our method is based on the direct prediction of dependencies between text spans, where the roles and central claim are derived automatically from the obtained dependencies through simple rules.

%%%%%%%%%%%%%%%%%%%%%%%%%%%%%%%%%%%%%%%%%
\paragraph{Biaffine Argument Parser.} 

Each task we presently address is a dependency tree construction task. The terminal nodes in the tree can be handcrafted ADUs or elementary discourse units predicted by a discourse parser. In the latter case, an additional structural function is introduced to combine several elementary DUs into one argumentative DU.  Given a sequence of $n$ discourse units $u_{1}, u_{2}, ... u_{n}$, elementary or argumentative, we first encode each discourse unit with CLS-pooling of a pretrained transformer into a vector $\mathbf{v_i} \in \mathbb{R}^{d_{LM}}$:
\begin{equation}
\mathbf{v_{i}} = \text{Encoder}(w_1w_2...w_k)
\end{equation}
and over the obtained representations run the biaffine dependency parsing model proposed by \citet{dozat2016deep}. In our model, the arc labels are argumentative functions, such as ``support'' or ``attack''. The central claim is encoded as an extra function, ``cc'', and it is the only function that is allowed to be assigned to the parentless node (the root).

The additional \texttt{root} node, which is a fictional parent of the real tree root, is randomly encoded into vector $\mathbf{v}_0$. The matrix $\mathbf{V} \in \mathbb{R}^{(n+1) \times d_{LM}} = [\mathbf{v}_0, \mathbf{v}_1, ..., \mathbf{v}_n]$ is then passed through four feedforward layers to get the parent-wise and dependent-wise arcs and functions hidden representations:

\begin{equation}
\begin{aligned}
H^{\text{(arc-parent)}} = \text{FF}^{\text{(arc-parent)}}(\mathbf{V}) \\
H^{\text{(arc-dep)}} = \text{FF}^{\text{(arc-dep)}}(\mathbf{V}) \\
H^{\text{(function-parent)}} = \text{FF}^{\text{(function-parent)}}(\mathbf{V}) \\
H^{\text{(function-dep)}} = \text{FF}^{\text{(function-dep)}}(\mathbf{V}) \\
\end{aligned}
\end{equation}
Those are used to score each possible parent for each dependent with bilinear attention:

\begin{dmath}
s_i^{\text{(arc)}} = H^{\text{(arc-parent)}} \mathbf{U} H^{\text{(arc-dep)}\top} + \mathbf{b^{\text{(arc)}}}
\end{dmath}
where $\mathbf{U}$ and $\mathbf{b^{\text{(arc)}}}$ are trainable. 

Due to the fact that a statement's role is directly related to its function towards its parent, roles are not predicted in a learnable way. Instead, roles are inferred directly from dependencies. Since the predicted central claim is the proponent's claim by definition, we traverse the predicted function-labeled dependency tree, assigning the role ($\text{``pro''} = \overline{\text{``opp''}}$) to the visiting node $i$ with parent $j$ as follows:
\begin{equation}
\text{role}_{i} = 
\begin{cases}
    \text{``pro''}, & \text{if function$_{i} = $ ``cc''}; \\
    \overline{\text{role}_j}, & \text{if function$_{ij} = $ ``attack''}, \\
    \text{role}_j, & \text{otherwise}.
    \end{cases}
\end{equation}
We examine performance on two main methods. \textbf{Biaffine Argument Parser (BAP)} uses a biaffine dependency parser as described above. \textbf{Discourse-driven Biaffine Argument Parser (DBAP)} additionally takes into account the discourse relations in a rhetorical tree.

%%%%%%%%%%%%%%%%%%%%%%%%%%%%%%%%%%%%%%%%%
\paragraph{Discourse-driven BAP.}

In order to incorporate rhetorical structures into argument parsing, we enhance the arc scores (3) with the corresponding discourse coefficients $\boldsymbol{C}^{\text{(RST)}}$ obtained from the rhetorical tree: $S^{\text{(arc)}} = S^{\text{(arc)}} \circ \boldsymbol{C}^{\text{(RST)}}$. The RST constituency trees are converted into RST dependencies.

In the simplest case, discourse coefficients are predicted from the $n\times n$ binary adjacency matrix $A^{\text{(RST-adj)}}$, where $a^{\text{(RST-adj)}}_{ij} = 1$ if there is a discourse relation going from discourse unit $i$ to the nucleus\footnote{In case of multiple nuclei, the leftmost nucleus.} DU $j$:

\begin{equation}
\boldsymbol{C}^{\text{(RST)}}  = \boldsymbol{\uptheta} A^{\text{(RST-adj)}} + \mathbf{b}^{\text{(RST)}}
\end{equation}
where $\boldsymbol{\uptheta}$ and $\mathbf{b}^{\text{(RST)}}$ are trainable scalar parameters controlling the effect of any discourse relation on the arc scores.

The type of rhetorical relation between the two DUs should also be considered when learning discourse coefficients, since some relations may not reflect the argumentative structure. This is accomplished by encoding the rhetorical label of each possible arc into a scalar value with an additional trainable layer. For this, we represent the labeled rhetorical dependency tree as the $n\times n \times k$ adjacency matrix  $A^{\text{(RST-full)}}$ with $a^{\text{(RST-full)}}_{ij}$ being one-hot encoded rhetorical relation going from discourse unit $i$ to the nucleus $j$. The discourse coefficients are then computed as

\begin{dmath}
\boldsymbol{c}_{ij}^{\text{(RST)}} = FF^{\text{(rst-arc)}}(a_{ij}^{\text{(RST-full)}}) = \sigma(a_{ij}^{\text{(RST-full)}\top}\boldsymbol{\Theta} + \mathbf{b}^{\text{(RST)}}),
\end{dmath}
where $\boldsymbol{\Theta}$ contains the trainable weights of specific rhetorical relations.  As an activation function $\sigma$ we use ReLU to prevent negative coefficients.

Finally, it is important to consider that for certain discourse relations the nuclearity-defined RST arc direction may contradict the direction of the argument. For this case, we also examine the inverted rhetorical relations:

\begin{dmath}
\boldsymbol{c}_{ij}^{\text{(RST)}} = FF^{\text{(rst-arc)}}(a_{ij}^{\text{(RST-full)}}) + FF^{\text{(rst-inv)}}(a_{ji}^{\text{(RST-full)}}).
\end{dmath}
Apart from penalizing predictions that contradict argumentative rhetorical relations, it also rewards inverting discourse relations that naturally oppose argument (e.g. \textsc{Preparation}).

\begin{table}
\small
\begin{tabular}{llp{0.31\textwidth}}
\toprule
Data  & \#  & Text                                                                                                                                               \\ \midrule
En    & 1      & \textcolor{blue}{Actually} it would be justified if all German universities charged tuition fees.                                                            \\
      & 2      & As long as it is \textcolor{blue}{ensured} that the funds really benefit the universities directly, \textcolor{blue}{one} can continue to regard this as social justice.              \\
      & 3      & \textcolor{blue}{Those who study later decide this early on, anyway.}                                                                                                 \\
      & 4      & \textcolor{blue}{It’s} always \textcolor{blue}{possible to take out} a student loan or \textcolor{blue}{to earn} a scholarship.                                                                           \\
      & 5      & To oblige \textcolor{blue}{non-academics to finance others' degrees through} taxes, however, \textcolor{blue}{is not just.}                                                            \\ 
\midrule
Ru$\rightarrow$En & 1      & \textcolor{blue}{In fact,} it would be justified if all German universities charged tuition fees.                                                            \\
      & 2      & As long as it is \textcolor{blue}{guaranteed} that the funds really benefit the universities directly, \textcolor{blue}{we} can continue to regard it as social justice.                \\
      & 3      & \textcolor{blue}{In any case, the question of further training must be decided in advance.}                                                                           \\
      & 4      & \textcolor{blue}{You can} always \textcolor{blue}{take} a student loan or \textcolor{blue}{get} a scholarship.                                                                                           \\
      & 5      & However, \textcolor{blue}{it is unfair} to oblige \textcolor{blue}{people who do not belong to scientific circles to pay for someone else's education by collecting additional} taxes. \\ \bottomrule
\end{tabular}
\caption{\label{tab:ru2en_example} Example of a text paraphrase by argumentative discourse unit (ADU), \texttt{micro\_k002}. }
\end{table}

\begin{figure}
\centering
    \includegraphics[width=0.35\textwidth]{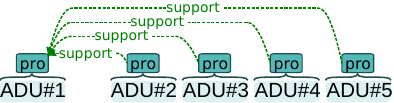}
    \caption{Example of the simplest argumentative structure, \texttt{micro\_k002}.}
    \label{fig:micro_k002}
    \par\bigskip\bigskip
    
    \begin{subfigure}[b]{0.47\textwidth}
        \centering
        \includegraphics[width=\textwidth]{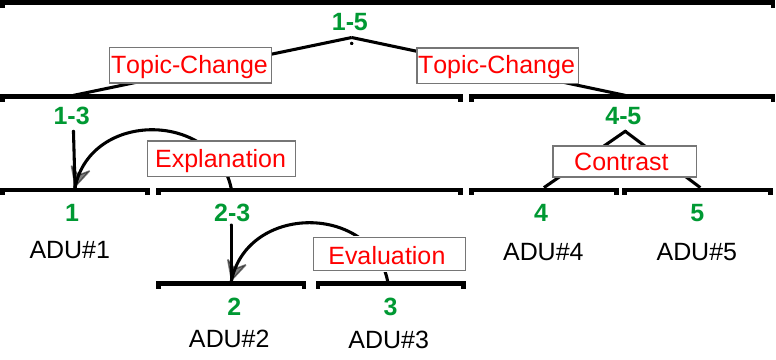}
        \caption{Original text in English (\texttt{En}).}
        \label{fig:rst_ex_en}
    \end{subfigure}
    \par\bigskip
    \begin{subfigure}[b]{0.47\textwidth}
        \centering
        \includegraphics[width=\textwidth]{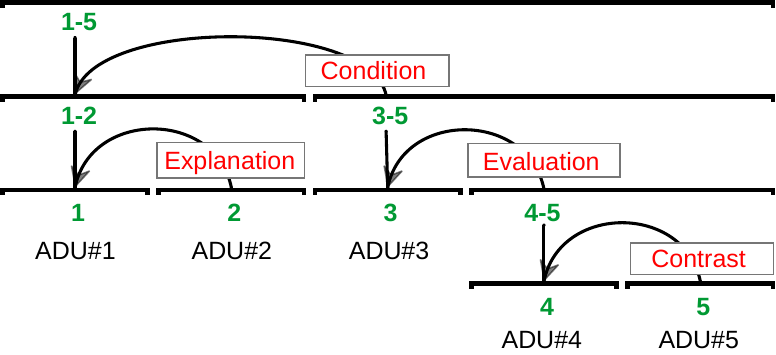}
        \caption{Paraphrase (\texttt{Ru$\rightarrow$En}).  }
        \label{fig:rst_ex_ru2en}
    \end{subfigure}
    \par\bigskip
    \begin{subfigure}[b]{0.47\textwidth}
        \centering
        \includegraphics[width=\textwidth]{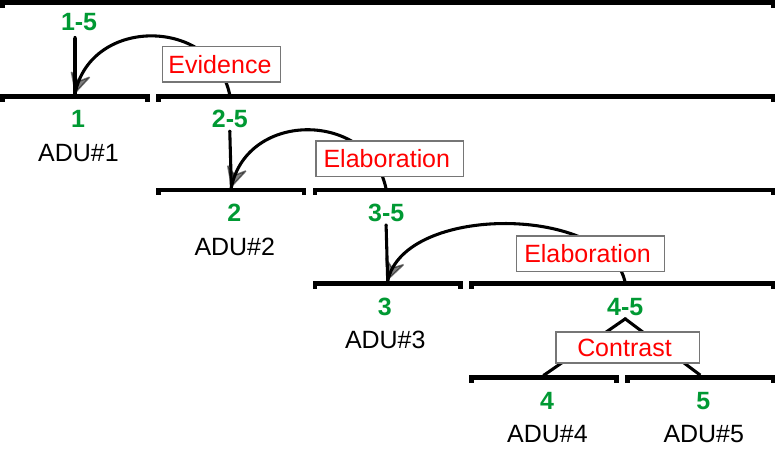}
        \caption{Original text in Russian (the exact source for the \texttt{Ru$\rightarrow$En} version).}
        \label{fig:rst_ex_ru}
    \end{subfigure}
    \par\bigskip
    \begin{subfigure}[b]{0.47\textwidth}
        \centering
        \includegraphics[width=\textwidth]{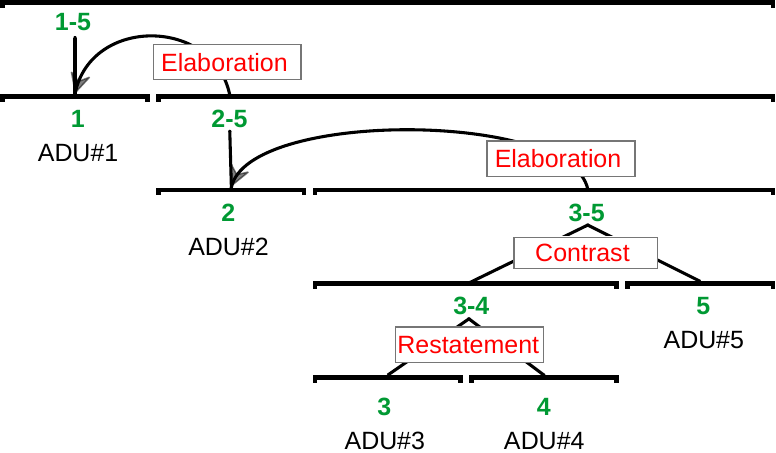}
        \caption{Paraphrase in Russian (literally follows the \texttt{En} version).  }
        \label{fig:rst_ex_en2ru}
    \end{subfigure}
    
    \caption{Four RST structure variants predicted for the document \texttt{micro\_k002} reduced to the relations between argumentative discourse units.} 
    %See Table~\ref{tab:ru2en_example} for the text references in English and \Cref{fig:micro_k002} for the argument structure annotation.}
    \label{fig:rst_examples}
\end{figure}

%%%%%%%%%%%%%%%%%%%%%%%%%%%%%%%%%%%%%%%%%
\section{Collecting the Structure Variations}
\label{sec:collecting_dv}

RST annotations are known to have a low intra-annotation agreement due to the ambiguity of discourse. Differences in annotation, magnified by the intrinsic limitations of statistical models in language understanding, lead to the unstable behavior of rhetorical analyzers. \textbf{To identify discourse variations, we use paraphrases of the annotated ADUs.}

%%%%%%%%%%%%%%%%%%%%%%%%%%%%%%%%%%%%%%%%%
\subsection{Semi-automated Back Translation} 
As pointed out by \citet{da2010comparing}, the use of translation strategies has a noticeable impact on rhetorical structures. In order to paraphrase, we use back translation over a parallel corpus of argumentative annotation.

In the Russian-language version of the Argumentative Microtexts, both parts of the original corpus have been manually translated from English into Russian by ADU \citep{fishcheva2019cross}. It is a literary translation and often does not correspond to the original in the number of clauses and sentences in ADU. Such paraphrases introduce pronounced differences in rhetorical structure from the original. They also cause an unstable parser to change its prediction most significantly. In order to get different retellings of the same argumentative structures, we additionally obtained the literal \textit{Human English~$\rightarrow$~Russian} and \textit{Human Russian~$\rightarrow$~English} machine translation preserving original ADU boundaries. We use the recent multilingual NLLB model \texttt{nllb-200-distilled-1.3B} \citep{costa2022no} for both directions, achieving 31.6\% BLEU in \textit{En~$\rightarrow$~Ru} and 29.2\% BLEU in \textit{Ru~$\rightarrow$~En} translation measured against handcrafted argumentative texts. 

Table~\ref{tab:ru2en_example} shows an example of the resulting paraphrase for a simple argumentative structure. 
In \Cref{fig:micro_k002}, it is shown that ADU \#2-5 support the central claim (\#1) independently. 
The semi-automated back translation helps to rephrase the individual statements within an argument slightly (ADU \#1, \#2, \#4) or significantly (ADU \#3, \#5).

\begin{table}[h]
\small
\center
\begin{tabular}{@{}lllll@{}}
\toprule
Lang        & Constituent & Nuclearity  & Relation    & Avg  \\ \midrule
En          & 0.56 ± 0.3  & 0.27 ± 0.5  & 0.35 ± 0.4  & 0.39 ± 0.3 \\ 
Ru          & 0.50 ± 0.3  & 0.26 ± 0.4  & 0.29 ± 0.4  & 0.35 ± 0.3 \\ 
\bottomrule
\end{tabular}
    \caption{The consistency of discourse parsing across different versions of a single text in the same language (mean ± std). The EDUs are reduced according to the gold ADU segmentation. Fleiss Kappa measures are computed following the \citet{iruskieta2015qualitative}’s method.}
    \label{tab:agreement}
\end{table}

%%%%%%%%%%%%%%%%%%%%%%%%%%%%%%%%%%%%%%%%%
\subsection{Analyzing Paraphrases from a Discourse Perspective}

In this study, we employ the recent end-to-end RST parsers for English\footnote{The models trained on RST-DT corpus.} \citep{zhang-etal-2021-adversarial} and Russian \citep{chistova2020rst}.

First, we assess the diversity of rhetorical structures, guided by the gold ADU segmentation in the corpus. 
Figure~\ref{fig:rst_examples} illustrates\footnote{rstWeb \citep{zeldes-2016-rstweb} is used to create all the RST visualizations in this paper.} variations of the rhetorical structure for the paraphrases obtained for the example in Table~\ref{tab:ru2en_example}, assuming that the leaves of the discourse tree are the annotated ADUs. None of the obtained RST trees matches the expert argument annotation (\Cref{fig:micro_k002}), although in each variant the most nuclear discourse unit in the RST tree (ADU\#1) naturally corresponds to the central claim in the argumentative structure.
A comparison using \citet{iruskieta2015qualitative}’s method reveals that two variants predicted by the same parser for English (\Cref{fig:rst_ex_en,fig:rst_ex_ru2en}) have Fleiss’ Kappa of 0.06 for nuclearity annotation and -0.04 for constituency annotation.
Nuclearity agreement for two variants predicted by the same parser for Russian (\Cref{fig:rst_ex_ru,fig:rst_ex_en2ru}) is 0.6 while constituency annotation agreement is 0.32. The agreement values are obtained with the RST-Tace tool proposed by \citet{wan-etal-2019-rst}. The original trees from Figure~\ref{fig:rst_examples} with EDUs intact are additionally shown in Appendix~\ref{sec:appendix_predicted_rs}, \Cref{fig:rst_examples_full}; these illustrate how RST structure varies within individual ADUs.

Table~\ref{tab:agreement} shows the pairwise Kappas for each language averaged over the corpus. Rarely, when an ADU does not entirely belong to an isolated RST discourse unit, its label is assigned to several DUs.  According to the results, Fleiss' Kappa values yield moderate agreement for unlabeled tree construction (Constituent) and fair agreement for nuclearity and relation assignments. Coherence of nuclearity, the feature directly related to identifying the central idea in the text, is the lowest on average.  Constituent Kappa equals 1.0 in 22\% of English and 18\% of Russian text pairs. The perfectly same rhetorical structure is found in 4\% of text pairs in English and 8\% in Russian. According to the results, the chosen strategy of paraphrasing helps to collect rhetorical structures with high variability.

\subsection{Training Data Augmentation}

In the experiments with both BAP and DBAP methods, the rephrased texts and the results of their discourse analysis are used for training data augmentation.

\begin{table*}[!ht]
\center
\small
\begin{tabular}{@{}lllllllll@{}}
\toprule
Data & Features                     & cc          & ro          & fu          & at          & UAS         & LAS         \\ \midrule
En   & All                          & 87.3 ± 6.2  & 74.4 ± 4.9  & 75.9 ± 6.4  & 50.7 ± 4.3  & 56.3 ± 5.2  & 50.1 ± 5.1  \\
En   & -BC                          & 85.9 ± 6.9  & 71.4 ± 5.7  & 75.0 ± 7.1  & 51.2 ± 4.2  & 56.3 ± 5.1  & 49.4 ± 5.1  \\
En   & -Cues                        & 88.4 ± 7.1  & 71.1 ± 6.4  & 73.0 ± 7.1  & 50.9 ± 3.6  & 56.8 ± 6.2  & 49.2 ± 6.6  \\
\midrule
En   & -BC, -Cues                   & 84.8 ± 6.3  & 71.9 ± 5.2  & 73.0 ± 5.9  & 51.4 ± 4.9  & 56.1 ± 5.9  & 48.5 ± 5.6  \\
Ru $\rightarrow$ En  & -BC, -Cues   & 82.0 ± 7.0  & 72.5 ± 6.7  & 72.7 ± 4.8  & 52.9 ± 3.3  & 56.5 ± 4.3  & 50.4 ± 4.3 \\
\midrule
Ru   & -BC, -Cues                   & 85.3 ± 3.9  & 73.4 ± 7.0  & 74.5 ± 3.6  & 56.5 ± 4.3  & 60.4 ± 4.7  & 52.9 ± 4.6  \\
En $\rightarrow$ Ru  & -BC, -Cues   & 87.3 ± 6.8  & 73.8 ± 6.4  & 73.8 ± 6.3  & 57.5 ± 4.4  & 61.8 ± 5.7  & 54.7 ± 5.5  \\
\bottomrule
\end{tabular}
    \caption{Performance of the baseline Evidence Graphs argument parser \citep{peldszus-stede-2015-joint} on the original and paraphrased data (gold segmentation).
    }
    \label{tab:eg_baseline}
\end{table*}

%%%%%%%%%%%%%%%%%%%%%%%%%%%%%%%%%%%%%%%%%
\section{Experiments}
\label{sec:experiments}

We collected additional versions of discourse structures and describe our experiments on the original and augmented training data. All the experiments are conducted on the first two of the ten 5-fold cross-validation splits from the experiments of \citet{peldszus-stede-2015-joint}. Since there was no validation data in the original splitting, we leave 15\% of the training data in each fold for validation. The training data is supplemented with the second crowd-sourced part of the corpus introduced by \citet{skeppstedt-etal-2018-less}. Following related work on the dataset \citeyearpar{peldszus-stede-2015-joint,peldszus-stede-2016-rhetorical,skeppstedt-etal-2018-less}, we use the simplified functions set, where ``support'', ``example'', and ``link'' functions are encoded as ``support'', while ``rebut'' and ``undercut'' are encoded as ``attack''. We leverage spaCy\footnote{\url{https://spacy.io/}. The models \texttt{en_core_web_lg} and \texttt{ru_core_news_lg}.} for feature extraction. 

All the experiments including pretrained language models were conducted with the \texttt{Microsoft/mDeBERTa\_v3} \citep{he2021debertav3}, the multilingual model sufficient for both languages.

\subsection{Experimental Setup}

Each model is trained on an NVIDIA Tesla V100 GPU. On average, it costs 25 seconds per epoch for parsing on gold segmentation and 39 seconds per epoch for end-to-end parsing with 30 to 75 training epochs total (on the original training sets; the augmentation doubles the training data).

The hyperparameters are tuned on the development subset of the corresponding split. Adam optimizer is used with a weight decay of 0.1 and a dropout rate of 0.2; $\beta = (0.9, 0.9)$. We use a learning rate of 2e-5 for the language model, while the randomly initialized layers have a learning rate of 2e-6. The discourse coefficients are trained with a learning rate of 2e-2. The dimension of the arc representation is 100 and the dimension of the tag representation is 50. The maximum sequence length is set to 150 tokens and the batch size is 4.

\subsection{Evaluation}

To evaluate the argument tree parsing, in addition to the attachment scores (UAS, LAS), we use the evaluation metrics introduced by \citet{peldszus-stede-2015-joint}. That is, we additionally report the macro-averaged F1 for central claim detection (\texttt{cc}), role assignment (\texttt{ro}), function tagging (\texttt{fu}), and F1 for positive attachment (\texttt{at}). To determine the statistical significance of pairwise comparisons, we perform paired t-test.

\begin{table*}[!ht]
\center
\small
\begin{tabular}{@{}lllllllll@{}}
\toprule 
Lang  & Method      & Augmented  & cc                & ro                & fu                & at                 & UAS                & LAS               \\ \midrule
En    & BAP         & No         & 88.3 ± 4.9        & \bftab71.1 ± 5.7  & 77.1 ± 4.6        & 53.8 ± 6.8         & 59.1 ± 6.8         & 52.9 ± 6.3        \\
      &             & Yes        & 88.9 ± 4.7        & 69.2 ± 3.9        & \bftab78.3 ± 4.9  & 56.2 ± 5.9         & 61.2 ± 5.8         & 55.1 ± 5.9        \\
      & DBAP        & No         & \bftab90.3 ± 3.3  & 68.8 ± 6.9        & 77.3 ± 3.2        & 59.7 ± 7.4*        & 64.5 ± 6.6*        & 56.2 ± 5.3*       \\
      &             & Yes        & 89.5 ± 4.3        & 68.8 ± 7.6        & 76.5 ± 3.1        & \bftab60.1 ± 4.3** & \bftab64.6 ± 4.1*  & \bftab56.6 ± 3.2* \\ 

\midrule

Ru    & BAP         & No         & \bftab90.5 ± 5.7  & 69.3 ± 7.8        & 78.9 ± 4.2        & 56.1 ± 6.3        & 61.7 ± 6.6         & 55.2 ± 6.7        \\
      &             & Yes        & 90.3 ± 2.8        & 66.9 ± 6.9        & 77.5 ± 4.3        & 56.1 ± 5.1        & 61.6 ± 4.7         & 53.9 ± 5.7        \\
      & DBAP        & No         & 90.3 ± 5.7        & 68.9 ± 2.5        & \bftab79.8 ± 3.6  & 59.8 ± 5.3        & \bftab64.6 ± 5.8   & \bftab58.0 ± 3.6  \\
      &             & Yes        & 88.3 ± 6.4*       & \bftab69.9 ± 5.4  & 77.2 ± 6.1        & \bftab60.6 ± 4.9* & \bftab64.6 ± 5.8   & 57.0 ± 5.8        \\ 
\bottomrule
\end{tabular}
    \caption{Performance of the biaffine argument parsers on the original and augmented  data (gold segmentation). Results that differ significantly from those of the non-augmented BAP are marked with * ($p < 0.05$) or ** ($p < 0.005$).
     }
    \label{tab:biaffine_gold_segm}
\end{table*}

\subsection{Baseline}

We  run  the  baseline MST model introduced by \citet{peldszus-stede-2015-joint}\footnote{\url{https://github.com/peldszus/evidencegraph}}. It predicts an argumentative dependency tree over the given discourse units from bags of words and bigrams, bags of discourse connectors and their associated relations, POS tags, punctuation, Brown clusters \citep{brown1992class} for words and bigrams, and occurrence of ADUs in the same sentence. Table \ref{tab:eg_baseline} shows the baseline results.  Due to the lack of discourse connectors vocabulary with annotated discourse relations for Russian, no markers or relations-related features were used in the multilingual experiments (\texttt{-Cues}). The same applies to Brown clusters, which are not available for Russian (\texttt{-BC}). Excluding these features from the original model for English results on average in a 2.5\% decrease in F1 for central unit detection and role assignment, 2.9\% for function classification, and a 1.6\% decrease in LAS. Regardless, excluding them is necessary to standardize experiments with multilingual data.

\paragraph{Does machine translation violate reasoning?} In Table~\ref{tab:eg_baseline}, we additionally report the results on paraphrases (\texttt{En$\rightarrow$Ru}, \texttt{Ru$\rightarrow$En}).
The results on the machine translations are marginally better than on the original handcrafted data, except for identifying the central claim in English data and functions in Russian. The F1 scores for role and function identification, however, do not represent the quality of argumentation tree construction because role and function classes are imbalanced. Attachment scores are higher on paraphrases. It seems that the reason for this is that every translation step simplifies the argumentative markers.
We conclude that the collected additional data is nearly as useful as the original.

\subsection{Segmentation}

\begin{figure}
\centering
    \includegraphics[width=0.48\textwidth]{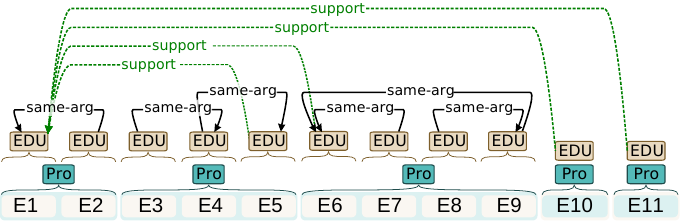}
    \caption{Argument tree representation in the end-to-end parser, \texttt{micro\_k002:En}. See \Cref{fig:rst_ex_en_full,fig:micro_k002} for reference.}
    \label{fig:segmentation_illustr}
\end{figure}

In end-to-end argument parsing, the elementary discourse units (EDUs) are considered leaves of the argument tree. Whenever an ADU matches a subtree of multiple EDUs, we preserve the discourse relations structure by assigning a ``same-arg'' argumentative function to every intra-ADU relation (\Cref{fig:segmentation_illustr}). Adding the third function class did not change the architecture of the model.

%%%%%%%%%%%%%%%%%%%%%%%%%%%%%%%%%%%%%%%%%
\section{Results and Discussion}
\label{sec:results}

\paragraph{Gold segmentation} The results are shown in Table~\ref{tab:biaffine_gold_segm}. The models with discourse perform significantly better than those without discourse, while the ones without discourse perform better than the baselines (Table~\ref{tab:eg_baseline}). Although performance increases with textual paraphrases of the same arguments, adding discourse structure variations over the same segmentation can hinder performance by introducing noise to the training data.

Appendix~\ref{sec:appendix_discourse_coefficients} presents the interpretation of the DBAP models trained on rhetorical structures.

\paragraph{Joint Segmentation and Parsing}

\begin{table*}[!ht]
\center
\small
\begin{tabular}{@{}lllllllll@{}}
\toprule 
Lang  & Method      & Augmented  & cc               & ro               & fu               & at               & UAS              & LAS              \\ \midrule
En    & BAP         & No         & \bftab86.8 ± 6.1 & 60.3 ± 4.2       & \bftab40.0 ± 2.7 & 39.2 ± 5.0       & 40.8 ± 8.0       & 23.1 ± 6.8       \\
      &             & Yes        & 86.3 ± 4.8       & \bftab64.5 ± 5.0*& 39.3 ± 3.0       & 42.0 ± 5.6       & 40.0 ± 6.7       & 25.5 ± 5.7       \\
      & DBAP        & No         & 85.8 ± 5.5       & 60.4 ± 6.2       & 39.3 ± 2.7       & 65.7 ± 2.9**     & 59.0 ± 4.7**     & 23.0 ± 5.2       \\
      &             & Yes        & 84.8 ± 5.0       & 62.9 ± 5.7       & 39.1 ± 2.8       & \bftab66.7 ± 4.2**& \bftab62.2 ± 3.8** & \bftab26.3 ± 7.1 \\
\midrule
Ru    & BAP         & No         & 88.6 ± 6.9       & 60.6 ± 5.8       & 42.3 ± 2.6       & 42.5 ± 6.4       & 45.4 ± 7.9         & 28.6 ± 6.4       \\
      &             & Yes        & \bftab90.2 ± 5.7 & 58.9 ± 3.5       & \bftab43.6 ± 2.5 & 43.5 ± 5.5       & 47.2 ± 8.0         & 30.7 ± 6.9       \\
      & DBAP        & No         & 86.5 ± 5.2       & 59.7 ± 5.5       & 42.9 ± 2.9       & 60.7 ± 4.9**     & \bftab59.6 ± 8.6** & \bftab31.7 ± 5.8 \\
      &             & Yes        & 87.5 ± 5.1       & \bftab60.7 ± 4.9 & 41.9 ± 3.5       & \bftab61.0 ± 3.8** & 58.2 ± 4.2**     & 29.9 ± 7.4       \\
\bottomrule
\end{tabular}
    \caption{Test results of the end-to-end biaffine argument parsers. Results that differ significantly from those of the non-augmented BAP are marked with * ($p < 0.05$) or ** ($p < 0.005$).
     }
    \label{tab:biaffine}
\end{table*}

Table \ref{tab:biaffine} shows the end-to-end parsing performance on the same test data when considering EDUs as terminal nodes. In order to compare the BAP and DBAP models consistently, the ``same-arg'' function is excluded from evaluation. While comparing BAP with DBAP in this setting is still not entirely fair, the BAP results can be viewed as a non-structural baseline.
When the second discourse structure variant is added, the training data provides variability in the representation of the connections between the same leaf nodes. It helps to find the general discourse patterns, resulting in a better performance on original test data in English. 
No improvement has been observed in Russian data, which highlights the differences between nuclearity interpretations in the two RST corpora. As a result of merging several relations into one, the nuclearity definition for some relations in the RST corpus for Russian differs from the original theoretical definition. In \textsc{Cause-effect} and \textsc{Purpose} relations, the nucleus always implies the logical effect, regardless of the author's intention. This affects the adequacy of the converted dependency discourse tree.

%%%%%%%%%%%%%%%%%%%%%%%%%%%%%%%%%%%%%%%%%
\section{Conclusion}
\label{sec:discussion}

In this paper, we constructed the first end-to-end argument parser on the Microtexts corpus. We show that using predicted rhetorical structures as initial data allows training a deep end-to-end model on the small corpus. We also report the results on gold segmentation as well as the interpretation of the obtained discourse coefficients. The proposed methods are evaluated in two languages; the first results on fully-fledged argument parsing for the Russian version of the corpus are reported. Our results suggest that argument mining can benefit from multiple variants of discourse structure.

%%%%%%%%%%%%%%%%%%%%%%%%%%%%%%%%%%%%%%%%%
\section*{Limitations}
\label{sec:limitations}
There are two limitations of this work. (1) The used corpus of Argumentative Microtexts contains only fully argumentative texts of moderate complexity. Real-world argument texts do not always consist only of argumentative statements. However, the method could potentially be used on other argumentation annotation corpora as well; one of the main reasons for choosing the corpus was to have a parallel full version in a second language. Another reason is the ability to match the EDU and ADU segmentations directly. (2) Although the amount of training data is artificially doubled, it may not be enough to train models on the proportionally increased noise.
We hope to investigate these directions in the future.

%%%%%%%%%%%%%%%%%%%%%%%%%%%%%%%%%%%%%%%%%
\section*{Acknowledgments}

The research was carried out using the infrastructure of the Shared Research Facilities «High Performance Computing and Big Data» (CKP «Informatics») of FRC CSC RAS (Moscow). This study was conducted within the framework of the scientific program of the National Center for Physics and Mathematics, section №9 ``Artificial intelligence and big data in technical, industrial, natural and social systems''.

%%%%%%%%%%%%%%%%%%%%%%%%%%%%%%%%%%%%%%%%%
\bibliographystyle{acl_natbib}
\bibliography{anthology,custom}

%%%%%%%%%%%%%%%%%%%%%%%%%%%%%%%%%%%%%%%%%
\newpage
\appendix
\section{Examples of RST Predictions}
\label{sec:appendix_predicted_rs}

\begin{figure*}
\centering
    \begin{subfigure}[b]{\textwidth}
        \centering
        \includegraphics[width=0.8\textwidth]{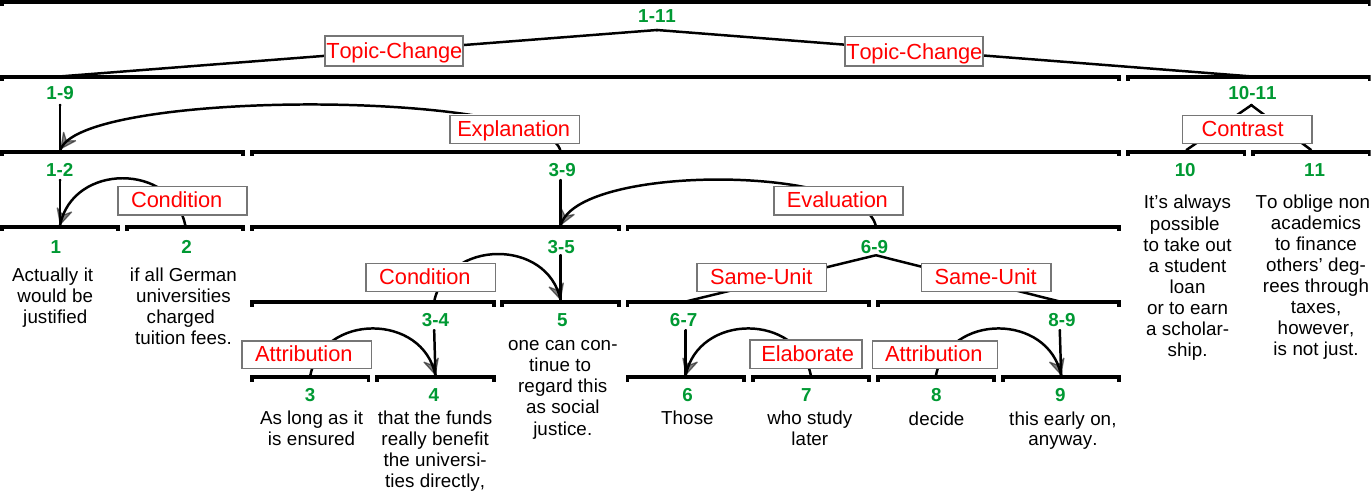}
        \caption{The original text in English (\texttt{En}).}
        \label{fig:rst_ex_en_full}
    \end{subfigure}
    \par\bigskip
    \par\bigskip
    \begin{subfigure}[b]{\textwidth}
        \centering
        \includegraphics[width=0.9\textwidth]{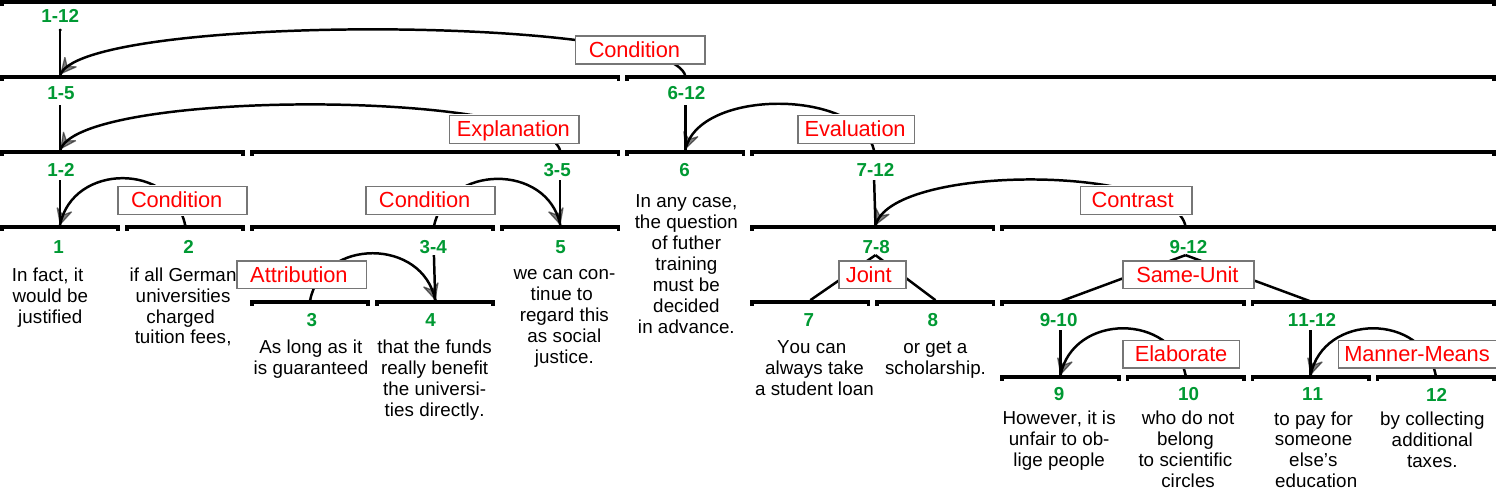}
        \caption{The paraphrase (\texttt{Ru$\rightarrow$En}).  }
        \label{fig:rst_ex_ru2en_full}
    \end{subfigure}
    \par\bigskip
    \par\bigskip
    \begin{subfigure}[b]{0.49\textwidth}
        \centering
        \includegraphics[width=\textwidth]{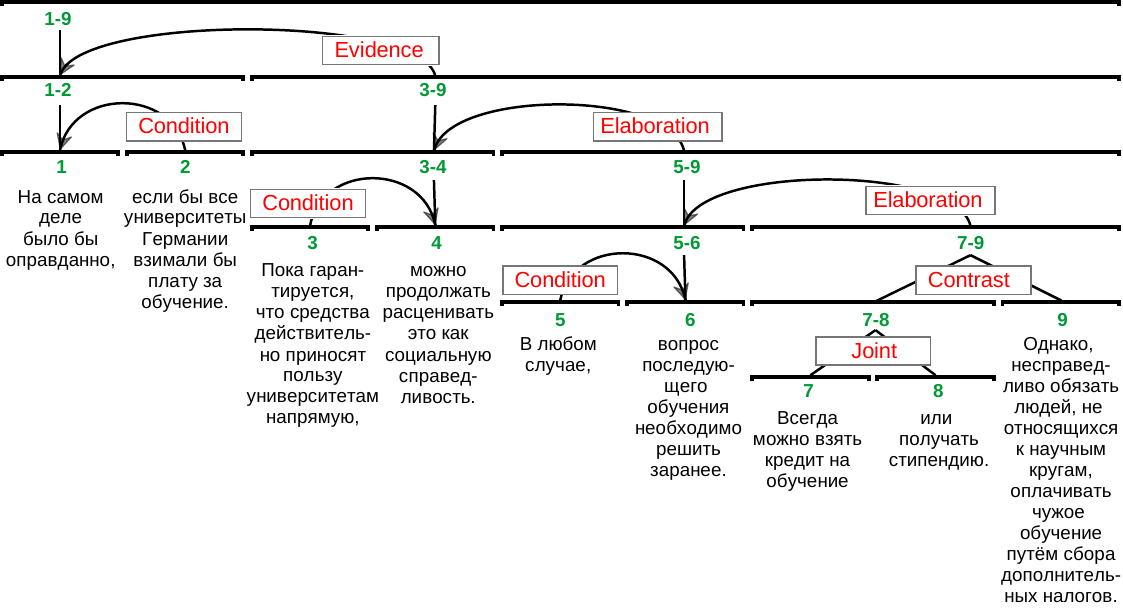}
        \caption{Original text in Russian (\texttt{Ru}): [In fact, it would be justified]$_1$ [if all German universities charged tuition fees.]$_2$ [As long as it is guaranteed that the funds really benefit the universities directly,]$_3$ [we can continue to regard this as social justice.]$_4$ [In any case,]$_5$ [the question of further training must be decided in advance.]$_6$ [You can always take a student loan]$_7$ [or get a scholarship.]$_8$ [However, it is unfair to oblige people who do not belong to scientific circles to pay for someone else's education by collecting additional taxes.]$_9$}
        \label{fig:rst_ex_ru_full}
    \end{subfigure}\hfill
    \begin{subfigure}[b]{0.46\textwidth}
        \centering
        \includegraphics[width=\textwidth]{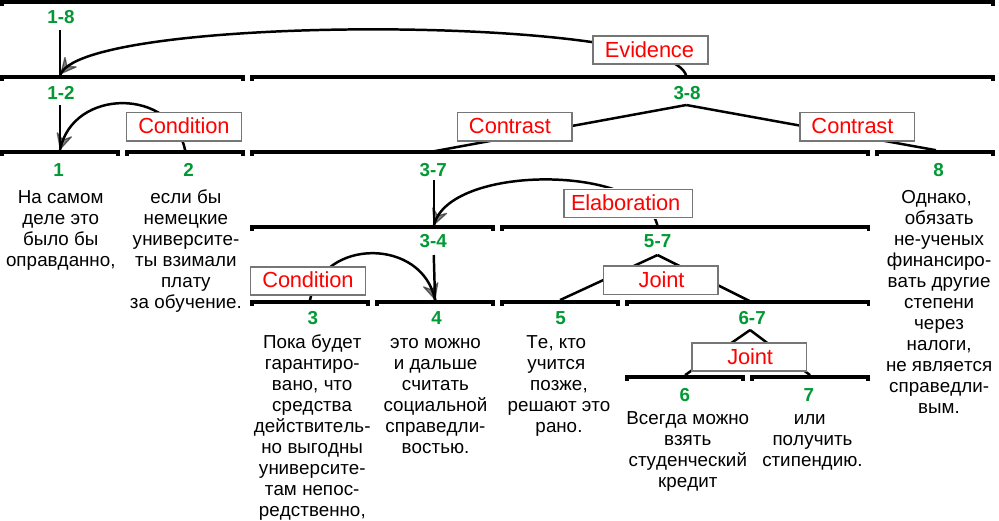}
        \caption{The paraphrase (\texttt{En$\rightarrow$Ru}): [Actually it would be justified]$_1$ [if all German universities charged tuition fees.]$_2$ [As long as it is ensured that the funds really benefit the universities directly,]$_3$ [one can continue to regard this as social justice.]$_4$ [Those who study later decide this early on, anyway.]$_5$ [It's always possible to take out a student loan]$_6$ [or to earn a scholarship.]$_7$ [To oblige non academics to finance others' degrees through taxes, however, is not just.]$_8$ }
        \label{fig:rst_ex_en2ru_full}
    \end{subfigure}
    \par\bigskip
    
    \caption{Four full discourse structure variants collected for the text \texttt{micro\_k002}.}
    \label{fig:rst_examples_full}
\end{figure*}

\Cref{fig:rst_examples_full} illustrates the predictions of the RST parsers with EDUs intact for four variants of a text example.

%%%%%%%%%%%%%%%%%%%%%%%%%%%%%%%%%%%

\section{Learned Discourse Coefficients by Relation}
\label{sec:appendix_discourse_coefficients}

\begin{figure*}[!ht]
\centering
    \begin{subfigure}[b]{0.9\textwidth}
        \centering
        \includegraphics[width=\textwidth]{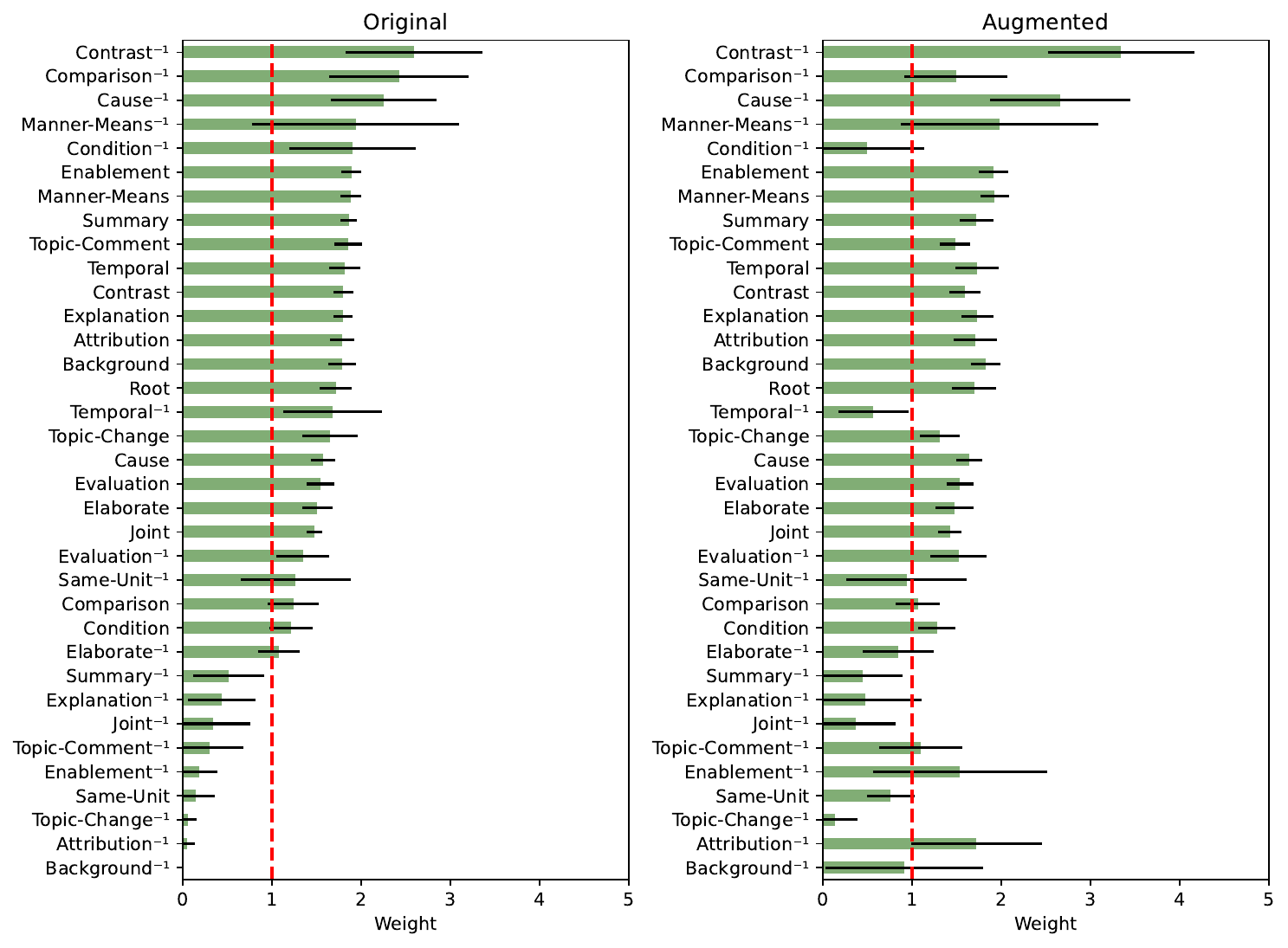}
        \caption{English.}
        \label{fig:c_rst_en}
    \end{subfigure}
    \par\bigskip
    \par\bigskip
    \begin{subfigure}[b]{0.9\textwidth}
        \centering
        \includegraphics[width=\textwidth]{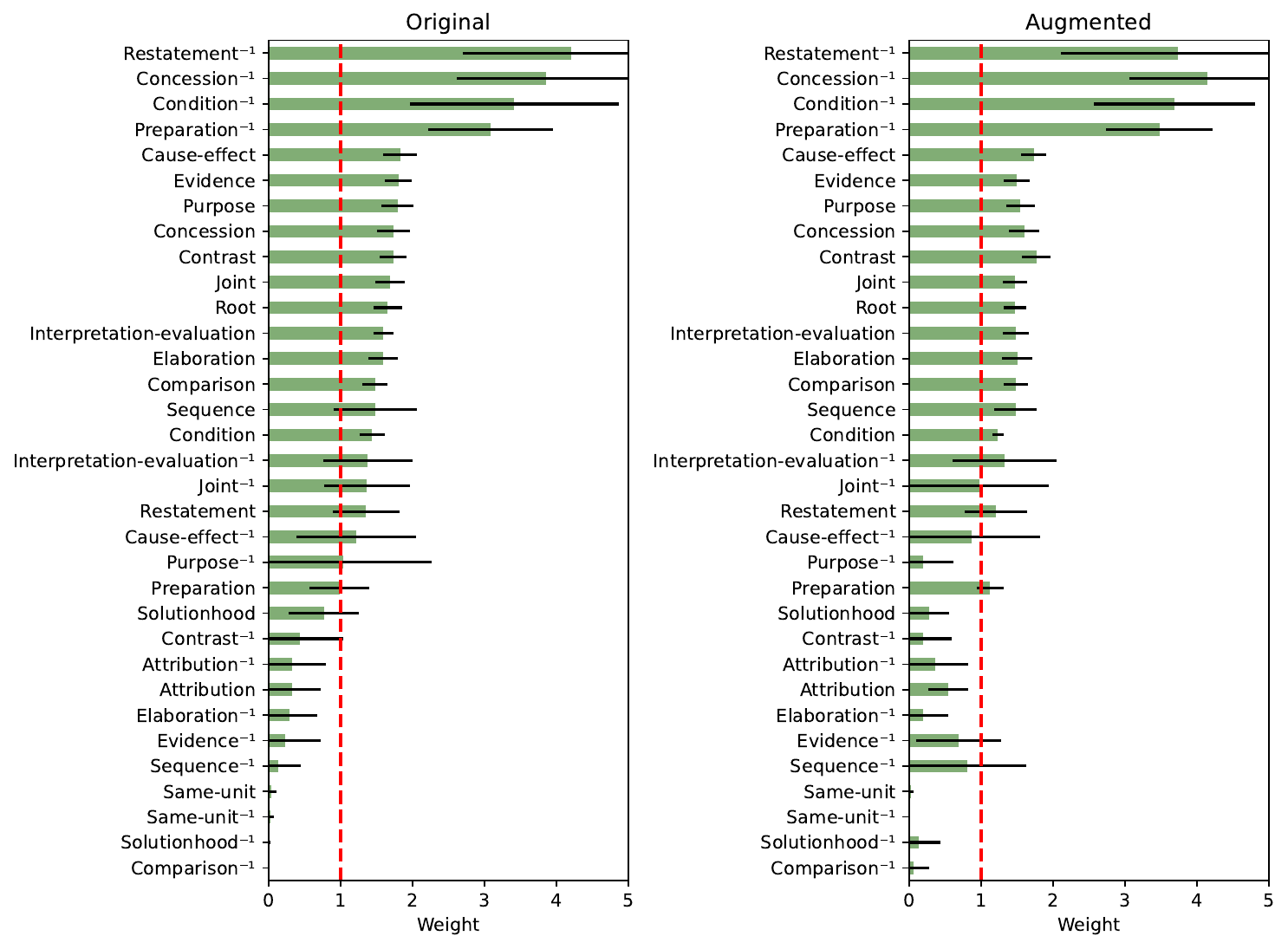}
        \caption{Russian.}
        \label{fig:c_rst_ru}
    \end{subfigure}
    \par\bigskip
    
    \caption{Statistics of the learned discourse coefficients $\boldsymbol{C}^{\text{(RST)}}$ for English (top) and Russian (bottom). }
    \label{fig:c_rst_ru_all}
\end{figure*}

We visualize the discourse coefficients $\boldsymbol{C}^{\text{(RST)}}$ in the trained DBAP models in \Cref{fig:c_rst_ru_all}. In light of the results, we divide rhetorical relations into four categories:

\subsection{The Argument's Companion}

The RST relations whose presence multiplies the likelihood of an argumentative function.

    \paragraph{For English\footnote{RST-DT benchmark relation set.}:}
    \begin{itemize}
        \item \textsc{Contrast}, \textsc{Contrast$^{-1}$}.         
        In the parsers trained on RST-DT, there are 17 coarse-grained relations which correspond to 78 different types of fine-grained RST relations. \textsl{Contrast}, \textsl{Concession} and \textsl{Antithesis} are treated by them as a single relation \textsc{Contrast}. 

        The simple \textsc{Contrast} cases are in full agreement with the argumentative structure:
        \begin{quote}
            {\small [Composting helps the natural environment]}
            $ \xleftarrow{\text{\small Attack}} $ 
            {\small [One drawback of composting is that not all material is beneficial to the environment.]}
            $_{\textsc{Contrast} \leftarrow}$
        \end{quote}
        
        The nuclearity and the argumentation may not agree if the fine-grained RST relation is something other than \textsl{Contrast}. The results for the \textsc{Contrast$^{-1}$} are consistent with earlier findings on the argumentativeness of the \textsl{Concession} relation, although the predicted nuclearity mostly opposes the argument direction:
        \begin{quote}
            {\small [It is true that social media is very beneficial for staying in contact with people far away,]}
            $_{\textsc{Contrast} \rightarrow} \xleftarrow{\text{\small Attack}} $ 
            {\small [but because there are no limits teens often wind up spending more time in that virtual world than in the real world.]}
        \end{quote}

        \item \textsc{Cause$^{-1}$}, \textsc{Cause}.        
        Despite the fact that causal relations directly reflect the argumentation, the predicted rhetorical nucleus often contradicts the direction of argument. Discourse-driven argument analysis can be hindered by this, especially if the discourse is deep and complex.
        \begin{quote}
            %$_{\texttt{micro\_k031}}$
            {\small [Pieces of dog poo on the pavements are a real danger.]}
            $ \xrightarrow{\text{\small Support}} $
            {\small [Increasing penalties is therefore the right way.]}
            $_{\textsc{Cause} \leftarrow}$
        \end{quote}
        When the RST nuclearity matches the causal logic, both relations are consistent with each other:
        \begin{quote}
            %$_{\texttt{micro\_k007}}$
            {\small [Supermarkets and shopping centres should be allowed to open any Sundays and holidays at their discretion.]}
            $ \xleftarrow{\text{\small Support}} $
            {\small [For in this way, Sunday shopping days would be better spread out through the year.]}
            $_{\textsc{Cause} \leftarrow}$
        \end{quote}

        \item \textsc{Enablement}, \textsc{Manner-Means}, \\\textsc{Explanation}.        
        The RST relations known for their innate argumentative nature. Often found within a sentence and featured by explicit markers:
        \begin{quote}
            %$_{\texttt{micro\_c072}}$
            {\small [Supermarkets should charge for plastic bags]}
            $ \xleftarrow{\text{\small Support}} $ 
            {\small [in order to encourage the use of reusable bags.]}
            $_{\textsc{Enablement} \leftarrow}$
        \end{quote}
        \begin{quote}
            %$_{\texttt{micro\_c072}}$
            {\small [Recycling helps the environment]}
            $ \xleftarrow{\text{\small Support}} $ 
            {\small [by keeping non-natural things out of it.]}
            $_{\textsc{Manner-Means} \leftarrow}$
        \end{quote}

        \item \textsc{Summary}.        
        As with the \textsc{Contrast}, the \textsc{Summary} relation class in parsers also embodies another fine-grained RST relation, \textsl{Restatement}. 
        In short texts, this relation also might be mistakenly assigned to examples of the similar implicit \textsc{Elaborate} and \textsc{Explanation}.
        \begin{quote}
            %{\texttt{micro\_c079}}
            {\small [Violent games cause people to react in an uncertain manner.]}
            $ \xleftarrow{\text{\small Support}} $
            {\small [It causes a type of stimulation that can trigger a violent reaction.]}
            $_{\textsc{Summary} \leftarrow}$
        \end{quote}

        \item \textsc{Temporal} (Rare, occurs in 1\% of data).        
        The RST parser fails to detect the causality in a sequence of events that the argument relation often assumes:
        \begin{quote}
            %{\texttt{micro\_c074}}
            {\small [Young children play violent games and assume this behavior is normal,]}
            $ \xleftarrow{\text{\small Support}} $ 
            {\small [and carry it out into the real world.]}
            $_{\textsc{Temporal} \leftarrow}$
        \end{quote}          

        \item \textsc{Joint}, \textsc{Elaborate}.        
        In \textsc{Elaborate}, a satellite (always on the right) provides additional detail for the state of affairs in a nucleus. In the multinuclear \textsc{Joint} nuclei are independent and equal in relation to the overall text function. Both relations are the most frequent in texts (38\% of parsed documents in English contain at least one \textsc{Elaborate} and 13\% at least one \textsc{Joint}). Rhetorical parsers are biased towards predicting these relations due to class imbalance in RST corpora.
        \begin{quote}
            %{\texttt{micro\_d12}}
            {\small [as it only has mediocre resolution]}
            $ \xleftarrow{\text{\small Support}} $
            {\small [and images in dark surroundings are often snowy.]}
            $_{\textsc{Joint} \leftarrow}$
        \end{quote}   

        \begin{quote}
            %{\texttt{micro\_c017}}
            {\small [Rhinos are becoming extinct.]}
            $ \xleftarrow{\text{\small Support}} $
            {\small [Poachers kill the rhinos with no regard.]}
            $_{\textsc{Elaborate} \leftarrow}$
        \end{quote}   
    \end{itemize}

    \paragraph{For Russian\footnote{RuRSTreebank relation set. The ADU texts in examples follow the English version of the Microtexts.}:}
    \begin{itemize}
        \item \textsc{Restatement$^{-1}$}.         
        One of the parts of this RST relation (nucleus or satellite) can act as a supporting argument for another in argumentation:
        
            \begin{quote}
                %{\texttt{micro\_c020}}
                {\small [They have allowed for easier communication,]} 
                $ \xrightarrow{\text{\small Support}} $ 
                {\small [so that means families are communicating when they otherwise wouldn't have.]}
                $_{\textsc{Restatement} \leftarrow}$ %\footnote{\foreignlanguage{Russian}{Они сделали общение доступнее, // это означает, что семьи общаются даже тогда, когда раньше общение было невозможно.}}
            \end{quote}   

            In English RST-DT parsing, the \textsl{Restatement} relation also exists as a part of the \textsc{Summary}.
    
        \item \textsc{Concession$^{-1}$}.         
        This class is described above as part of the \textsc{Contrast} relation in the English RST.
            \begin{quote}
                %{\texttt{micro\_c171}}
                {\small [Although this behavior may seem to hinder the child's independence and self reliance,]}
                $_{\textsc{Concession} \rightarrow}$ 
                $ \xleftarrow{\text{\small Attack}} $ 
                {\small [the child also has the knowledge that their parent will always be there for them.]}
            \end{quote}
    
        \item \textsc{Condition$^{-1}$}.         
        Parsing confusion occurs when this relation is expressed by ``even if''  {\fontencoding{T2A}\selectfont (``даже если'')}:
            \begin{quote}
                %{\texttt{micro\_b059}}
                {\small [Even if one might think that additional rent control is needed besides the current tenant protection,]}
                $_{\textsc{Condition} \rightarrow}$ 
                $ \xleftarrow{\text{\small Attack}} $ 
                {\small [one should not deny longstanding owners the opportunity to adjust their returns to market level.]}
            \end{quote}
    
        \item \textsc{Preparation$^{-1}$}.         
        Absent as a distinct relation in the RST-DT, \textsc{Preparation} can be viewed as the direct opposite of \textsc{Elaboration}. In this relation, the satellite sets or introduces a topic for the nucleus, yet contains only minimal information itself. The RST parser for Russian tends to assign this relation to the first statement in a text:
            \begin{quote}
                %{\texttt{micro\_c123}}
                {\small [Yes nuclear energy is safe.]}
                $_{\textsc{Preparation} \rightarrow}$ 
                $ \xleftarrow{\text{\small Support}} $ 
                {\small [There are many safeguards in place at power plants to prevent accidents.]}
            \end{quote}
    
        \item \textsc{Cause-effect}.         
        While semantically equivalent to the \textsc{Cause}/\textsc{Cause$^{-1}$} relations discussed above, in RST parsers for Russian the nuclearity of causal relations is determined by the logic of the described events, rather than the author's perspective. It is true that such rhetorical relations fit better with the logic of argumentation; however, converting to a rhetorical dependency tree in this circumstances can disrupt the whole text's logical coherence.
        \begin{quote}
            %{\texttt{micro\_c004}}
            {\small [As long as restraint and practicality are applied to hunting, the environment will not suffer.]}
            $_{\textsc{Cause-effect} \rightarrow}$
            $ \xrightarrow{\text{\small Support}} $ 
            {\small [Hunting is good.]}
        \end{quote}
        
        \item \textsc{Contrast}.        
        Similar to previous \textsc{Contrast} and \textsc{Contrast$^{-1}$}. In Russian RST, the \textsc{Contrast} relation is always multinuclear. Therefore, when converting to dependency, the head span is considered to be the statement on the left. The results show that this definition of \textsc{Contrast} is consistent with argumentative functions, while backward (left-to-right) arcs (\textsc{Contrast$^{-1}$}) are consistently penalized (see examples for English).
    
        \item \textsc{Purpose}. Corresponds to a part of the previously discussed \textsc{Enablement} in English RST parsing.
        \item \textsc{Evidence}. Part of the RST-DT \textsc{Explanation}.
        \item \textsc{Sequence}. Part of the \textsc{Temporal} in English RST. The description and an example for the sequential \textsc{Temporal} are given above.
        \item \textsc{Joint} and \textsc{Elaboration}. See \textsc{Joint}, \textsc{Elaborate}.
    \end{itemize}

    \subsection{Opposing Argument}
    The rhetorical relations consistently penalizing the argument arc probability.
    
    \paragraph{For English:} \textsc{Summary$^{-1}$}, \textsc{Joint$^{-1}$}, \textsc{Topic-Change$^{-1}$}, \textsc{Same-Unit}. 
    \paragraph{For Russian:} the arcs opposing argumentation (\textsc{Purpose$^{-1}$}, \textsc{Contrast$^{-1}$}, \textsc{Elaboration$^{-1}$}, \textsc{Joint$^{-1}$}), and non-argumentative relations (\textsc{Solutionhood}, \textsc{Solutionhood$^{-1}$}, \textsc{Attribution}, \textsc{Attribution$^{-1}$}, \textsc{Same-unit}, \textsc{Same-unit$^{-1}$}).

    \subsection{Vaguely Correlated}
    The average coefficient value merely exceeds one with a high deviation. The deviation shows that these rhetorical relations are often mispredicted in argumentative texts.
    
    \textbf{En:}    
    \textsc{Comparison$^{-1}$}, \textsc{Manner-Means$^{-1}$}, \textsc{Solutionhood$^{-1}$}, \textsc{Enablement$^{-1}$}, \textsc{Attribution$^{-1}$}.
    \textbf{Ru:}
    \textsc{Interpretation-evaluation$^{-1}$}.
    
    \subsection{Vaguely opposed}
    Values near or below one with a high deviation. 
    
    \textbf{En:} \textsc{Condition$^{-1}$}, \textsc{Temporal$^{-1}$}, \textsc{Same-Unit$^{-1}$}, \textsc{Explanation$^{-1}$}, \textsc{Background$^{-1}$}, \textsc{Topic-Comment$^{-1}$}.
    \textbf{Ru:} \textsc{Cause-effect$^{-1}$}, \textsc{Evidence$^{-1}$}, \textsc{Sequence$^{-1}$}.

%%%%%%%%%%%%%%%%%%%%%%%%%%%%%%%%%%%%%%%%%%
\section{License}

The licenses of the models, software and data used in this paper are listed below:
\begin{itemize}
        \item Argumentative Microtexts corpus \cite{peldszus2015annotated}: \textit{CC BY-NC-SA 4.0}.
        \item RST parser for English \citep{zhang-etal-2021-adversarial}, AllenNLP \citep{gardner-etal-2018-allennlp}: \textit{Apache License 2.0}.
        \item RST parser for Russian \citep{chistova2020rst}, RST-Tace \citep{wan-etal-2019-rst}, rstWeb \citep{zeldes-2016-rstweb}, Multilingual DeBERTa v3 \citep{he2021debertav3}, spaCy \citep{honnibal2020spacy}, Evidence Graph framework \citep{peldszus-stede-2015-joint}: \textit{MIT License}.
        \item NLLB MT model \citep{costa2022no}: \textit{CC-BY-NC 4.0}.
	% \item Argumentative Microtexts corpus \cite{peldszus2015annotated}: CC BY-NC-SA 4.0.
	% \item RST parser for English \citep{zhang-etal-2021-adversarial}: Apache License 2.0.
	% \item RST parser for Russian \citep{chistova2020rst}: MIT License.
 %        \item RST-Tace \citep{wan-etal-2019-rst}: MIT License.
	% \item rstWeb \citep{zeldes-2016-rstweb}: MIT License.        
	% \item NLLB MT model \citep{costa2022no}: CC-BY-NC 4.0.
 %        \item AllenNLP \citep{gardner-etal-2018-allennlp}: Apache License 2.0.
	% \item Multilingual DeBERTa v3 \citep{he2021debertav3}: MIT License.
 %        \item spaCy \citep{honnibal2020spacy}: MIT License.
	% \item Evidence Graph framework \citep{peldszus-stede-2015-joint}: MIT License.
\end{itemize}

\end{document}